\documentclass{article}

\usepackage{arxiv}

\usepackage[utf8]{inputenc}
\usepackage[T1]{fontenc}    
\usepackage{hyperref}      
\usepackage{url}           
\usepackage{booktabs}      
\usepackage{amsfonts}      
\usepackage{nicefrac}       
\usepackage{microtype}   
\usepackage{lipsum}

\usepackage{cite}
\usepackage{amsmath,amssymb,amsfonts}
\usepackage{algorithmic}
\usepackage{graphicx}
\usepackage{textcomp}
\def\BibTeX{{\rm B\kern-.05em{\sc i\kern-.025em b}\kern-.08em
    T\kern-.1667em\lower.7ex\hbox{E}\kern-.125emX}}

\usepackage[utf8]{inputenc}
\usepackage{array}
\usepackage[free-standing-units=true]{siunitx}
\usepackage{gensymb}   
\usepackage{svg}
\usepackage[export]{adjustbox}
\usepackage{hhline}
\usepackage{makecell, multirow} 
\usepackage{tabularx,booktabs}
\usepackage{lipsum}
\usepackage{ragged2e}
\usepackage[flushleft]{threeparttable}
\usepackage{pifont}
\newcolumntype{P}[1]{>{\centering\arraybackslash}p{#1}}
\newcolumntype{M}[1]{>{\centering\arraybackslash}m{#1}}
\newcolumntype{C}{>{\centering\arraybackslash}X} 
\newcommand{\cmark}{\ding{51}}
\newcommand{\xmark}{\ding{55}}

\newcommand{\RNum}[1]{\uppercase\expandafter{\romannumeral #1\relax}}
\newlength{\xfigwd}
\title{ElderSim: A Synthetic Data Generation Platform for Human Action Recognition in Eldercare Applications}

\author{
  Hochul Hwang\\
  Artificial Intelligence \& Robotics Institute\\
  Korea Institute of Science and Technology (KIST)\\
  Seoul 02792, Republic of Korea\\
  \texttt{hchlhwang@kist.re.kr} \\
   \And
 Cheongjae Jang \\
  Artificial Intelligence \& Robotics Institute\\
  Korea Institute of Science and Technology (KIST)\\
  Seoul 02792, Republic of Korea\\
  \texttt{jchastro@gmail.com} \\
     \And
 Geonwoo Park \\
  Artificial Intelligence \& Robotics Institute\\
  Korea Institute of Science and Technology (KIST)\\
  Seoul 02792, Republic of Korea\\
  \texttt{gunwop@naver.com} \\
   \And
 Junghyun Cho\\
  Artificial Intelligence \& Robotics Institute\\
  Korea Institute of Science and Technology (KIST)\\
  Seoul 02792, Republic of Korea\\
  \texttt{jhcho@kist.re.kr} \\
   \And
 Ig-Jae Kim \\
  Artificial Intelligence \& Robotics Institute\\
  Korea Institute of Science and Technology (KIST)\\
  Seoul 02792, Republic of Korea\\
  \texttt{drjay@kist.re.kr} \\
}

\begin{document}
\maketitle

\begin{abstract}
To train deep learning models for vision-based action recognition of elders' daily activities, we need large-scale activity datasets acquired under various daily living environments and conditions. However, most public datasets used in human action recognition either differ from or have limited coverage of elders' activities in many aspects, making it challenging to recognize elders' daily activities well by only utilizing existing datasets. Recently, such limitations of available datasets have actively been compensated by generating synthetic data from realistic simulation environments and using those data to train deep learning models. In this paper, based on these ideas we develop ElderSim, an action simulation platform that can generate synthetic data on elders' daily activities. For 55 kinds of frequent daily activities of the elders, ElderSim generates realistic motions of synthetic characters with various adjustable data-generating options, and provides different output modalities including RGB videos, two- and three-dimensional skeleton trajectories. We then generate KIST SynADL, a large-scale synthetic dataset of elders' activities of daily living, from ElderSim and use the data in addition to real datasets to train three state-of-the-art human action recognition models. From the experiments following several newly proposed scenarios that assume different real and synthetic dataset configurations for training, we observe a noticeable performance improvement by augmenting our synthetic data. We also offer guidance with insights for the effective utilization of synthetic data to help recognize elders' daily activities.
\end{abstract}


\section{Introduction}
\label{sec:intro}
    The need and importance of vision-based human action recognition (HAR) are growing in a wide range of eldercare services [1], including care robots [2], [3], smart surveillance [4], and health monitoring [5], [6]. Recently, the performance of vision-based HAR has been dramatically improved by deep learning methods [7]--[17], which require large-scale training datasets for accurate action recognition [7], [19]--[24] as mentioned in [18]. Accordingly, to train deep learning models to recognize elders' activities of daily living (ADL), we need large-scale datasets which contain activities acquired under various environments and conditions that we encounter in daily life.

    However, most public datasets used in HAR, including the NTU RGB+D dataset [21], which is frequently used as a benchmark, either differ from or have limited coverage of elders' daily activities in many aspects. Even if they have a large number of samples with various action classes, only a few action classes of such datasets match elders' ADL. Moreover, they are usually acquired from laboratory environments that deviate from the places of daily living. The way or speed of actions may also differ from the elders’ since they mostly consist of relatively young subjects’ actions. These differences can induce inaccurate action recognition results when a model trained on such datasets is tested on data of elders’ ADL [24].

    Recently, some datasets of elders' ADL have been publicly available [24], [25]. However, due to the limited data acquisition conditions, they often lack diversity in aspects such as background, camera viewpoint, and lighting condition. The low variations in a dataset can cause overfitting of deep learning models, especially for RGB-based HAR methods that are sensitive to the conditions above. An overfitted model will not generalize well and result in low recognition accuracy when applied to data obtained under conditions significantly different from the training datasets.

    A naive approach to this problem is to build a dataset that reflects all the conditions that arise in diverse real-world household environments. However, it is expensive and laborious to acquire such a dataset because of the combinatorial explosion [26], i.e., the number of data becomes exponentially large. More difficulties exist when acquiring labeled real data. Along with the expense of camera hardware, viewpoints are often restricted due to spatial limitations such as small-sized bathrooms or complex indoor environments. Personal privacy issues and physical limitations of the elders also make it more challenging to obtain a large-scale training dataset of good quality.

    To compensate for the limitations of available datasets and the difficulties of acquiring real data, recent studies endeavor to generate automatically-labeled synthetic data from virtual environments [27]--[29], [51]--[53] and further use those data to train deep learning models and enhance action recognition performance [29], [52]. In such virtual environments, we can freely adjust aspects such as backgrounds, subjects (or synthetic characters), camera viewpoints, and lighting conditions. Therefore, it becomes possible to customize the dataset that contains a large number of realistic data as needed. If such synthetic data are appropriately utilized for training deep learning models, we can expect those data to help fill the holes that reside in real-world datasets, e.g., the limited coverage in camera viewpoints and lighting conditions or the severe gap from the target data in subjects’ ages and backgrounds.

    In this paper, based on the above ideas we develop ElderSim, an action simulation platform that can generate synthetic data on elders' daily activities. We visualize the daily living environment and the characters of ElderSim to be as close as possible to those of the real-world using a recent three-dimensional rendering and modeling software. Considering the actual application to eldercare services, we model movements for 55 kinds of frequent daily activities of the elders, and offer variabilities in data acquiring options such as camera viewpoints and lighting conditions that change over time, to name a few. To summarize, ElderSim generates realistic daily living activities of synthetic characters with several adjustable data-generating options and provides different output modalities including RGB videos, two-dimensional (2D) and three-dimensional (3D) skeleton trajectories to further increase applicability. As an illustrating dataset generated from ElderSim, we release KIST SynADL, a large-scale simulated synthetic dataset of elders' activities. ElderSim and the KIST SynADL dataset is publicly available in \url{https://ai4robot.github.io/ElderSim}.

\begin{figure*}
\includegraphics[width=\linewidth]{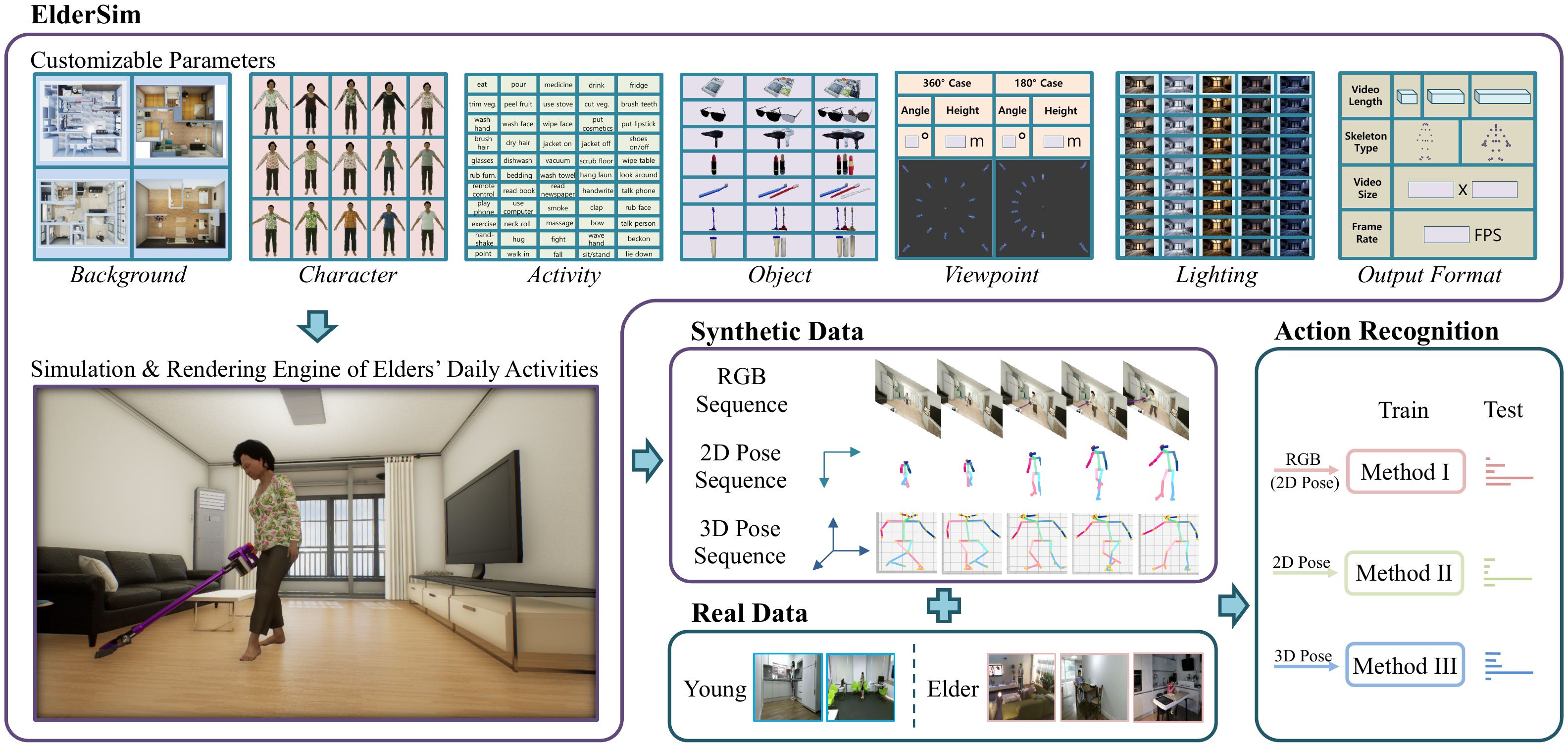}
\caption{Synthetic data generation process of ElderSim and the main pipeline of our proposed work. ElderSim generates synthetic RGB video, 2D, and 3D skeleton data based on the data-generating options that are customized by the user. Here, we experimentally augment synthetic data on real ones and train three different action recognition methods (Method \RNum{1}: Glimpse [11], Method \RNum{2}: ST-GCN [12], Method \RNum{3}: VA-CNN [13]) to scrutinize the effects of our generated data.}
\label{fig:eldersim_dev}
\end{figure*}

    We use KIST SynADL in addition to real datasets to train state-of-the-art HAR models, and validate the effectiveness of augmenting synthetic data. Unlike previous data augmentation studies focusing primarily on some limited benchmark datasets and experimental scenarios, we propose several new scenarios to examine various aspects which arise from recognizing elders' ADL. Specifically, in addition to cross-view and cross-subject train/test splits, widely considered in the literature, we newly introduce cross-age and cross-dataset splits that assume the real and synthetic training datasets of different configurations. Here, the two settings are focused more on the application to the elders. We also examine synthetic data augmentation for each of the three data modalities provided, namely RGB video, 2D and 3D skeleton. From the extensive experiments held with three action recognition architectures on two different real-world datasets, we show that augmenting our synthetic data for training increases recognition performance for most of the considered methods in various settings. We also offer some guidance with insights on utilizing synthetic data to help recognize elders' daily activities effectively. These points are of great importance since they can be easily combined with additional improvements in both deep learning models and objective functions for training to gain enhanced action recognition performance. The main pipeline of our work is described in Fig.~\ref{fig:eldersim_dev}.

    The paper is organized as follows. Section~\ref{sec:related_Work} presents related works, and we elaborate on ElderSim along with the synthetic dataset generated from ElderSim in Section~\ref{sec:elsersim_dev}. Section~\ref{sec:experiments} presents action recognition experiments augmenting our synthetic data. We conclude in Section~\ref{sec:conclusion}.

\section{Related Work}
\label{sec:related_Work}
    In this section, we introduce several HAR methods with various human activity datasets utilized in the literature. We also mention previous studies that exploit synthetic data to improve action recognition.

\subsection{Human Action Recognition}
Considering the temporal dimension along with the spatial dimension is essential for video understanding. Conventionally, these features were extracted by hand-crafted descriptors, including Histogram of Oriented Gradients (HOG) [30], Motion Boundary Histogram (MBH) [31], and Histograms of Optical Flow (HOF) [32], which is followed by a classifier such as Support Vector Machines (SVMs) for classification. Other methods used dense trajectories that track densely-sampled feature points and extract appearance and motion information with the previously mentioned descriptors along the trajectories [33], [34].

    With the development of powerful computation hardware and large-scale activity datasets, deep learning methods achieved profound performance in action recognition. As being one of the convolutional neural network (CNN)-based methods [7]--[9], [35], [36], [35] computes spatial information from a still frame and samples temporal motion from multiple-frame dense optical flow. The information is then inserted into a two-stream network that consists of a spatial and temporal CNN for better training. [36] first introduced 3D convolution for action recognition, which extracts features from both the spatial and temporal dimensions with a 3D kernel. [7] tested varying architectures on their own dataset, Sports-1M, showing that the Slow Fusion model outperforms other structures with different connectivity in time and also proposed an architectural method using lowered-resolution inputs to speed up training without any performance loss. [8] built a C3D (Convolutional 3D) network with 3D kernels and empirically showed the optimal kernel size and network architecture for improved action recognition performance in large-scale video datasets. Unlike [8], [9] inflated the filters and pooling kernels of deep 2D image classification models (e.g., Inception-v1 with batch normalization [37]) into 3D to gain from the advantages of ImageNet pre-training. They marked up performance with an additional optical-flow stream which is trained separately and tested with averaged-predictions. Without the support of multimodal inputs, [10] computes a single modality of raw RGB to a two-stream design by passing each pathway with a different frame rate. Spatial meanings are captured through one stream with a higher frame rate while the temporal information is learned through the other stream with a lower frame rate. The architecture in [11] learns to predict the attention windows in the feature space; extracted from a global model. A set of recurrent architectures are used to track the unstructured windows and classify actions from RGB video inputs.

    Skeleton-based HAR methods [12]--[16] have been studied to avoid various interferences of RGB appearance while using simpler data that are coordinates of several joints and their derived forms. These methods mainly obtain the human skeletal structure utilizing depth sensors [38] or human pose estimation algorithms [39], [40]. [16] uses the main LSTM network with a spatial attention module and a temporal attention module that holds different attention levels to select discriminative joint inputs and frame outputs, respectively. The three networks are jointly trained for optimization as an end-to-end training method. More recently, initiated by [12], graph convolutional network-based action recognition studies tried to understand the skeletal information as a graph and extract features using CNNs. [14] represented the tree-based natural human body structure as a directed acyclic graph for a better interpretation. They also adopt the two-stream method by feeding a graph that contains information of joints and bones to one network and another graph that contains the motion of joints and deformation of bones to the other network. [15] used temporally different-sized kernels instead of fixed ones as in [12] and added an additional spatial graph convolution layer branch to form a parallel structure. They further improved performance by using six modalities of input features including relative positions of joints and bones. Other multimodal fusion methods enhance performance by handling data of different domains, such as RGB and 3D skeleton [17].

\subsection{Real-World Activity Datasets}
    Diverse real-world human activity datasets have been publicly available with the emerging importance of robust human action recognition. Initial human activity datasets were relatively small-scale, having a small number of subjects and activity categories, until the early 2010s [41]--[44]. KTH [41], being one of the earliest databases, contains a single RGB modality of six action categories with simple motions such as \textit{walk}, \textit{run}, and \textit{clap}. Depth map information was firstly provided with RGB in MSR Action3D [42], which focused on game console interaction-based motions, including \textit{draw circle}, \textit{forward kick}, \textit{tennis swing}. As an extension of [42], MSR Daily Activity3D [43] covers living room daily activities, most containing human-object interaction captured by a Kinect sensor. RGBD HuDaAct [44] also deals with 12 daily activities of 30 students with RGB and depth modalities maintaining under 1,200 samples. Most of the early datasets contain only a few thousands of video samples and under 20 class categories, which allowed studies for hand-crafted methods without the support of deep learning. 
    Large-scale activity datasets emerged along with the advance of data-hungry action recognition methods [7], [19]--[24]. The initial version of Kinetics [19], obtained from YouTube, included 400 activity classes having more than 300K in total. The dataset is now extended to contain 700 classes with approximately 650K video clips. PKU-MMD [20] provides untrimmed daily activity video sequences in four modalities of RGB, depth, infrared (IR), and skeleton for the research field of action detection. A more recent multimodal dataset, MMAct [23], was released with seven modalities: RGB, skeleton, acceleration, and other sensor signals. NTU RGB+D [21] and its updated version [22] are extensively used as a benchmark dataset in recent human action recognition literature. They respectively captured 60 and 120 action categories, including daily actions, medical situations, and human-human interactions. Both versions are provided with RGB, depth, 3D skeletons, and IR data, having almost 115K samples for the updated version.
    
    Some datasets are acquired under more varied settings to better reflect the conditions that are likely to occur in real-world applications. Multi-view human activity datasets were obtained from various camera viewpoints by simultaneously using several cameras or changing the camera viewpoints in a different trial [21], [22], [45], [46]. UESTC [46] considered human-robot interaction (HRI) applications for action recognition from arbitrary viewpoints. The dataset includes eight fixed viewpoints with arbitrary viewpoints sampled from the entire 360° horizontal directions. There also exist other datasets that target action recognition for specific applications, such as eldercare [24], [25]. ETRI-Activity3D [24] captured elders’ ADL from several viewpoints considering mobile robots’ heights for care robotic services. Toyota Smarthome [25] is another dataset on elders' ADL that possesses severe class imbalance and intraclass variation by capturing unscripted daily activity videos of the elderly. 
\subsection{Synthetic Data Exploitation}
    To provide abundant training data for deep learning methods to avoid overfitting, some studies focused on utilizing synthetic data. Synthetic data generation is considered cost-effective and customizable since users can manipulate data reflecting one’s needs without any additional data capturing middleware or subjects. Some studies generate synthetic data using generative adversarial networks (GANs) [47], [48] or composite methods based on existing real data [49], [50]. Another group of studies uses computer graphics and game engine techniques to simulate data and exploit them for deep learning tasks [27]--[29], [51], [52].

    [48] uses two adversarial generative networks to train instance-level pairwise cross-view connection knowledge and performs robust action recognition with additional training data generated for deficient views. [49] composites realistic images to overcome the laborious manual labeling process for the 3D skeleton, depth, and motion. The human motion is extracted based on motion capture (MoCap) recordings, randomized textures, viewpoints, and lighting conditions are added on top of a static real-world background image to generate data; such data are applied to human depth estimation and human part segmentation tasks. The subsequent study [50] extracts 3D human dynamics using a 3D human shape estimation method and synthesizes other randomized components to render complementary training data to improve the action recognition from unseen viewpoints. 

    Here, we focus on game engine techniques to synthesize data without any reference RGB data and generate realistic videos by considering various factors, including the context of the background, physics, and object interaction. Various game engine-based data generation studies were conducted in fields where data acquisition in various environments is highly expensive, such as autonomous systems [51], [52] and robotics [27], [28]. For human action recognition, Souza et al. [29] initially generated abundant synthetic training data under a variety of conditions with the Unity® game engine and enhanced action recognition performance by training with a mixture of real-world and generated data. However, most activity categories are not indoor activities of daily living hence not applicable to train models for eldercare applications. [53] introduced a simulation platform, developed in Unreal Engine 4® (UE4), to procedurally produce photorealistic synthetic videos of household activities in various modalities, but fails to provide details of the synthetic data augmentation effect in action recognition. [52] developed a simulation framework to automatically generate annotated training data from a game engine. They show outstanding action recognition accuracy in classifying five activities by training a shallow skeleton-based action recognition algorithm with their generated data. In this paper, we further explore the benefits of training synthetic data based on three state-of-the-art deep action recognition algorithms (fed with different data modalities containing RGB, 2D, and 3D skeleton) to classify 55 action classes. 
    
\section{ElderSim Development}
\label{sec:elsersim_dev}
    We now elaborate on how our elders' activity simulation platform, denoted as ElderSim, has been developed in detail. In the development, we focus on the following two aspects: 1) to visualize the virtual environment as close as possible to the real-world and 2) to reflect various situations that could be captured in real applications. To fulfill our first aim, we utilize a real-time photorealistic rendering platform Unreal Engine 4® (UE4) and a three-dimensional (3D) computer animation and modeling software called Autodesk Maya® (Maya). Using the two software, we construct the simulation environment of elders' daily living that resembles the real household backgrounds. We then model appearances and movements of synthetic characters based on the motion capture (MoCap) data obtained from the elders. To achieve the second objective, we consider 55 activity classes that sufficiently include the most frequent ADL of the elders. We also make it available to customize various camera viewpoints and lighting conditions, regarding care robot and smart surveillance applications. The following sections explain further development details and distinctive features of ElderSim. We then introduce KIST SynADL, a large-scale synthetic dataset generated from ElderSim.  
    
\subsection{Background}
\label{sec:background}
\begin{figure}
\centering \includegraphics[width=0.4\linewidth]{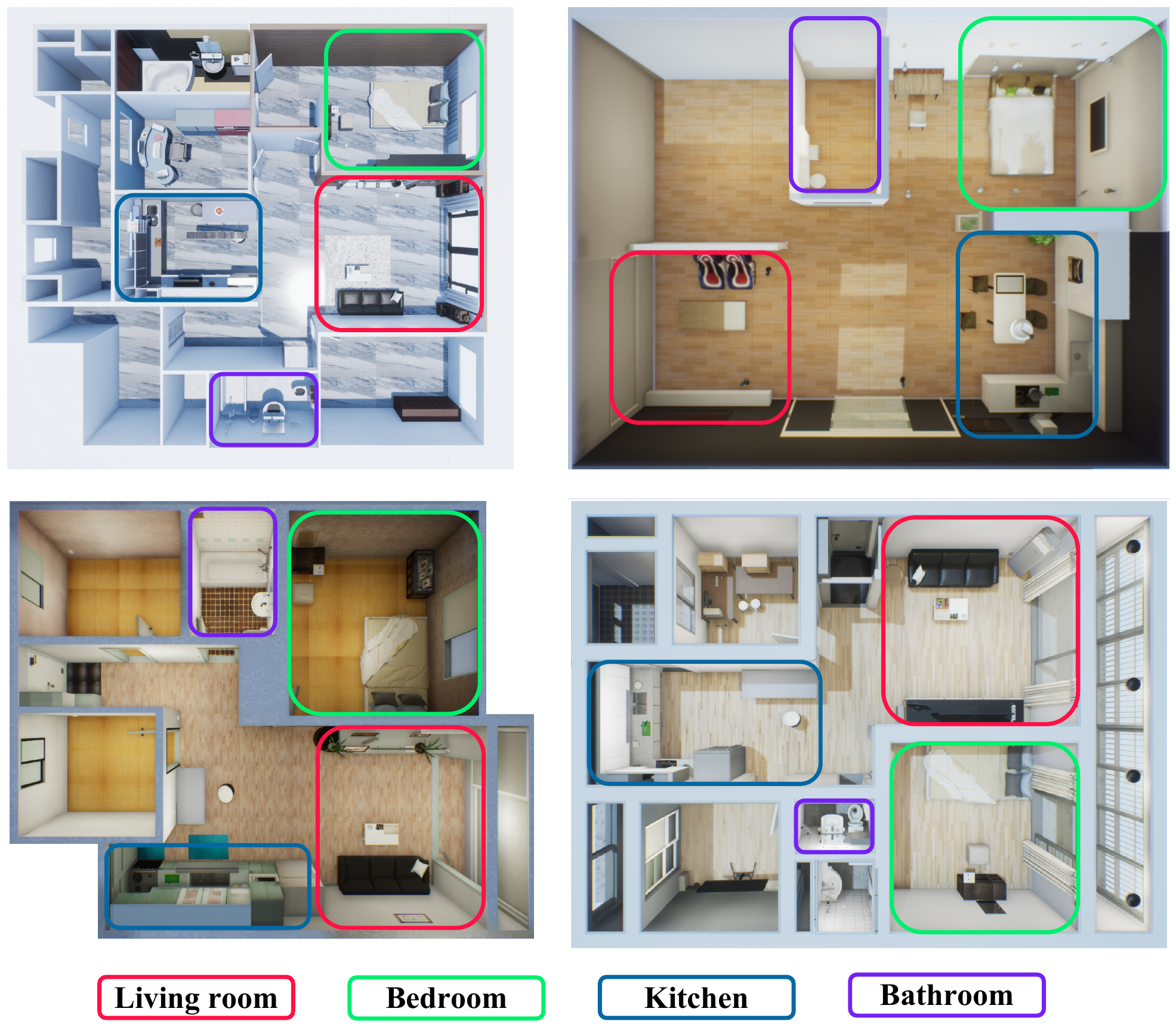}
\caption{The top view of four different residential household backgrounds implemented in ElderSim. Each household consists of four areas where daily activities are frequently occurred.}
\label{fig:background}
\end{figure}

    To provide realistic simulation backgrounds for elders’ daily living in ElderSim, we have modeled four residential houses based on their indoor measurements and photographs. House models can be added if necessary. When implementing the house models in ElderSim, the household background has become visually more realistic by using physics-based materials and the Post-Process Volume function in UE4. Each of the four house models contains four areas (living room, bedroom, kitchen, and bathroom) as shown in Fig.~\ref{fig:background}. In each area, we only simulate activities that are plausible to be performed (e.g., \textit{wash face} is simulated only in the bathroom while \textit{play with a mobile phone} is simulated in all four areas).

\subsection{Character}
\label{sec:character}
\begin{figure}
\centering\includegraphics[width=0.5\linewidth]{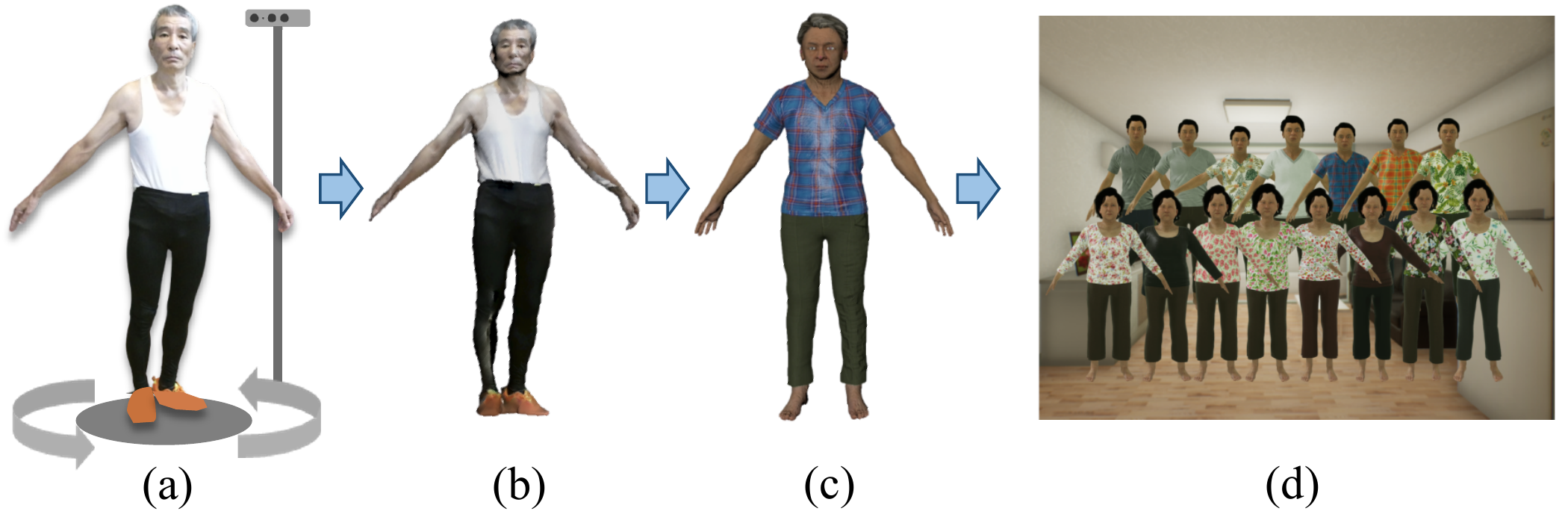}
\caption{Body shapes are captured from a depth sensor in the real-world (a). Depth information is used to model the synthetic character (b) and appropriate textures are applied (c) to reflect the various appearances (d).}
\label{fig:character}
\end{figure}

    We have modeled synthetic characters that imitate thirteen elder subjects (seven females and six males with average age and standard deviation as 69.92 and 3.36, respectively) and two relatively young subjects (a female and a male) in ElderSim. These subjects have been recruited to sufficiently represent a variety of body shapes and appearances. Their body shapes have been captured from Kinect depth sensors and utilized to design the body shape of synthetic characters in Maya. The faces of characters have been randomly created due to legal issues on portrait rights. In addition, different age-appropriate clothes have been applied to each character to enhance their appearance diversity. As a result, ElderSim can generate action data from fifteen synthetic characters possessing individual face, body shape, and appearance (see Fig.~\ref{fig:character}). The number of synthetic characters can be increased by implementing body shape transformation techniques in computer graphics, which are left for future work.

\subsection{Motion} 

    Following [24], we provide motions for 55 activity classes considered to be the most frequent ADL of the elders in ElderSim. To generate realistic motions for these activities, we utilize MoCap data obtained from the subjects recruited in Section~\ref{sec:character}. Sixteen digital MoCap cameras have captured the subject motions using 40 markers attached to the subjects' body. When acquiring data, there have not been any specific instructions for subjects to perform ADL to increase realism and diversity of motions. The obtained MoCap data have been rigged in Maya, i.e., skeletal templates and their movements that best fit the data are constructed. From the rigged data, the motions for synthetic characters are generated by adjusting the template’s kinematic parameters to those of each character and playing the constructed movements. To provide motion data in 3D skeleton modality, we define skeleton joints by attaching the sockets of UE4 to each character's 25 joints following two types of joint format labels used in OpenPose [40] and Kinect v2. Two-dimensional (2D) joint motions are obtained by projecting those in 3D to the image plane of each camera viewpoint using a transformation function in UE4.
    
\subsection{Viewpoint} 
\label{sec:viewpoint}
\begin{figure}
    \centering\begin{tabularx}{0.6\linewidth}{lCC}
        \small\textbf{360°} &
        \includegraphics[height=0.65\linewidth, width=0.65\linewidth]{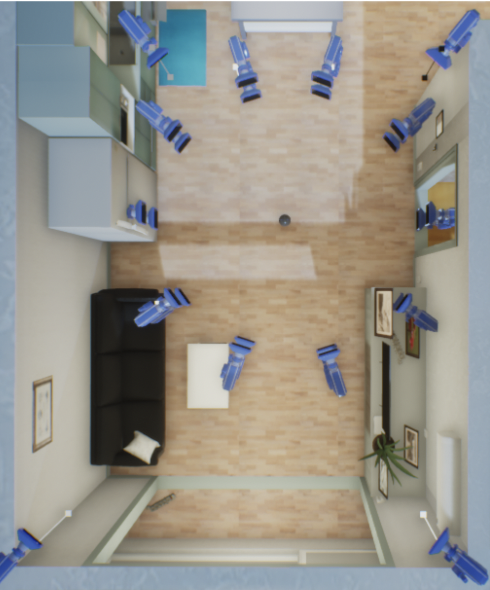} &
        \includegraphics[height=0.65\linewidth, width=\linewidth]{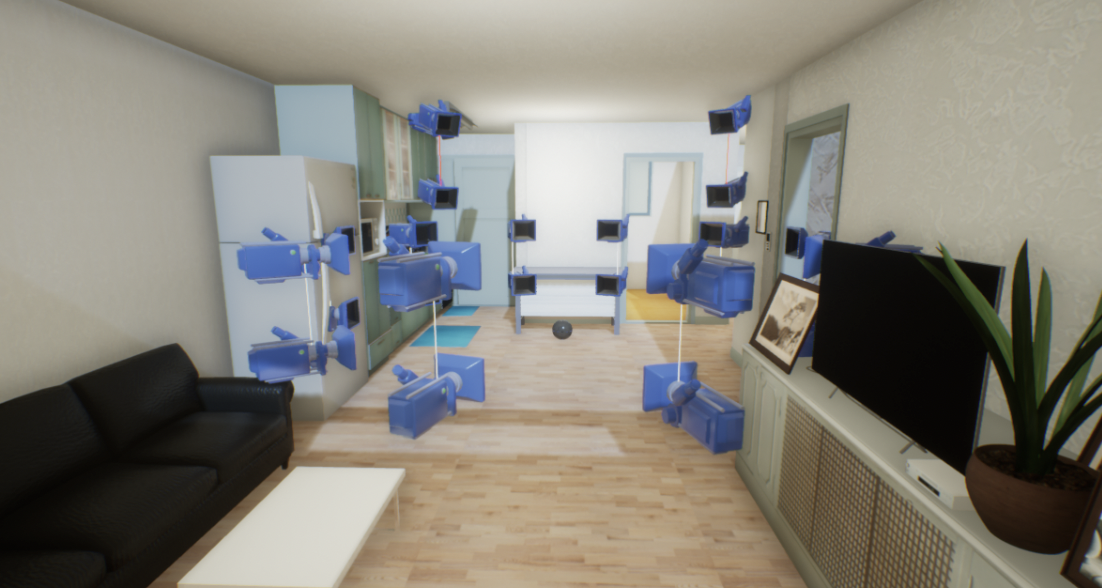}  \\
        \small\textbf{180°} &
        \includegraphics[height=0.65\linewidth, width=0.65\linewidth]{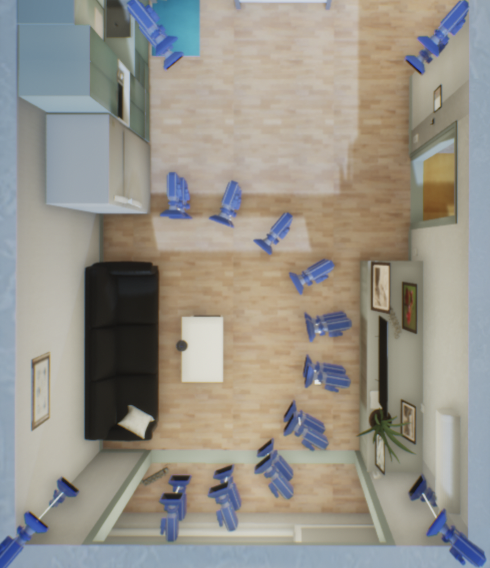} &
        \includegraphics[height=0.65\linewidth, width=\linewidth]{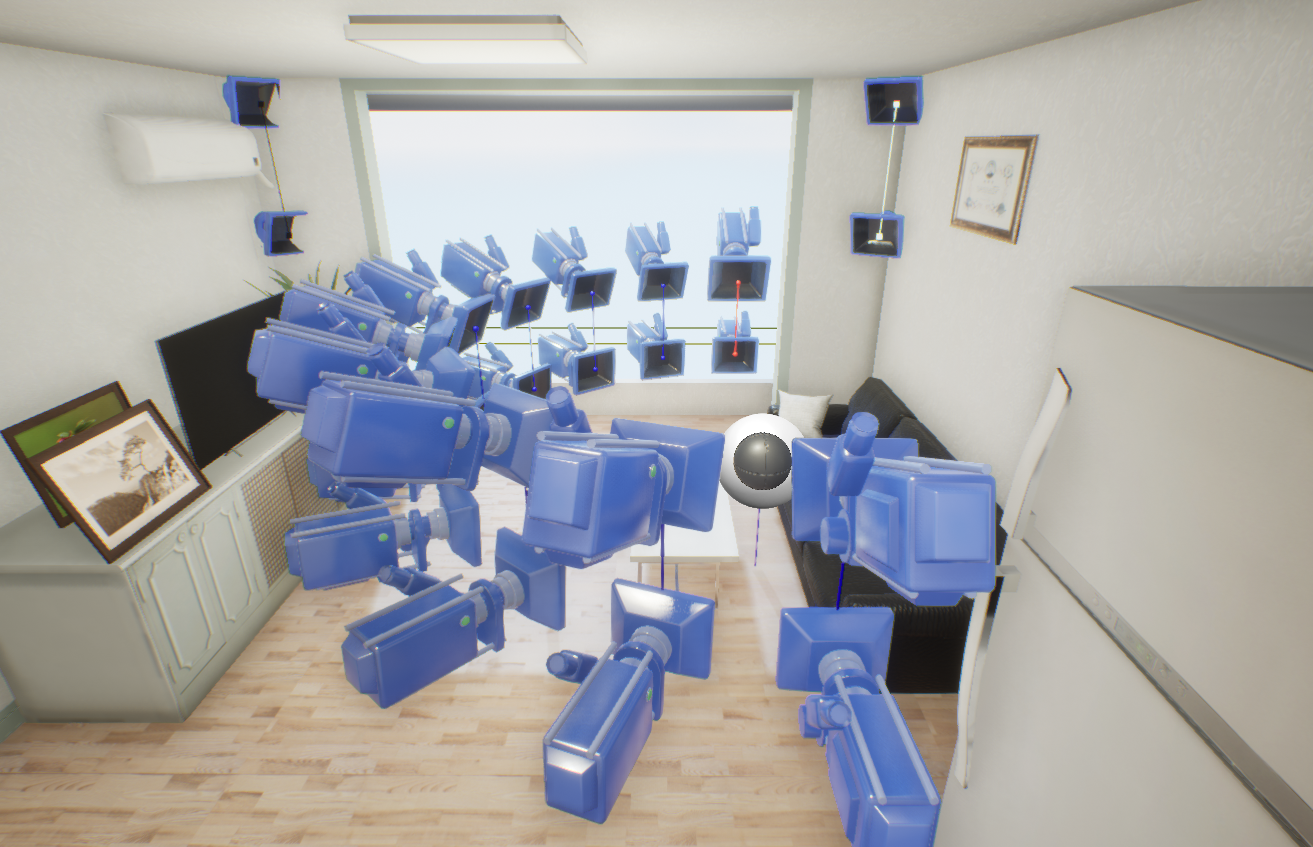}\\
        &\textbf{\small{Top view}} &\textbf{\small{Front view}}   \\
    \end{tabularx}
\caption{An example of the virtual camera setups based on the available viewpoints. Top and front views are chosen for a better interpretation. Here, 20 cameras represent robot-viewpoints and eight cameras on each corner of the room represent surveillance-viewpoints. The user can easily manage the number of viewpoints by adjusting the angle interval and a set of heights.} 
\label{fig:viewpoints}
\end{figure}

    Camera viewpoints in ElderSim contain robot- and surveillance-viewpoints, considering eldercare applications. Robot-viewpoints simulate video acquisition from care robots, and corresponding cameras are located on a circle to surround a target character with the circle radius appropriately defined from the range of the character's motion. The cameras have equal spacing on the circle with an angle interval $\phi$ and located at heights specified by a set of $h$ height values  $\delta=\{ \delta_1,\delta_2, ..., \delta_h \}$, where $\delta_i$ denotes the $i$-th height value. The angle interval $\phi$ and the set of height values $\delta$ are set to be user-adjustable parameters. Given these parameter values, the number of viewpoints ($V_{circle}$) can be expressed as 
    \begin{equation}
        V_{circle}=h\times floor(360^{\circ}/\phi).
    \end{equation}
    Such a circular camera layout may not be available occasionally due to obstacles in some backgrounds (e.g., when the character is sitting on a sofa, a wall behind the sofa hinders the rear robot-viewpoints); we then form viewpoints to cover a semicircle instead of a circle (see Fig.~\ref{fig:viewpoints}). In this semicircular camera layout, the number of viewpoints is given as
    \begin{equation}
        V_{semicircle}=h\times (floor(180^{\circ}/\phi)+1).
    \end{equation}
    To implement these camera layouts for robot-viewpoints in ElderSim, we define UE4 splines that contain multiple cameras vertically and position these splines according to the parameter settings. Meanwhile, surveillance-viewpoints simulate video acquisition from surveillance cameras such as closed-circuit televisions (CCTV). They are located at the height of \SI{1.5}{\metre} and \SI{2.2}{\metre} in four corners of each area to reflect the realistic camera installation, hence resulting in eight surveillance-viewpoints.

\subsection{Lighting} 
\begin{figure}
    \centering\includegraphics[width=0.5\linewidth, height=\linewidth, keepaspectratio]{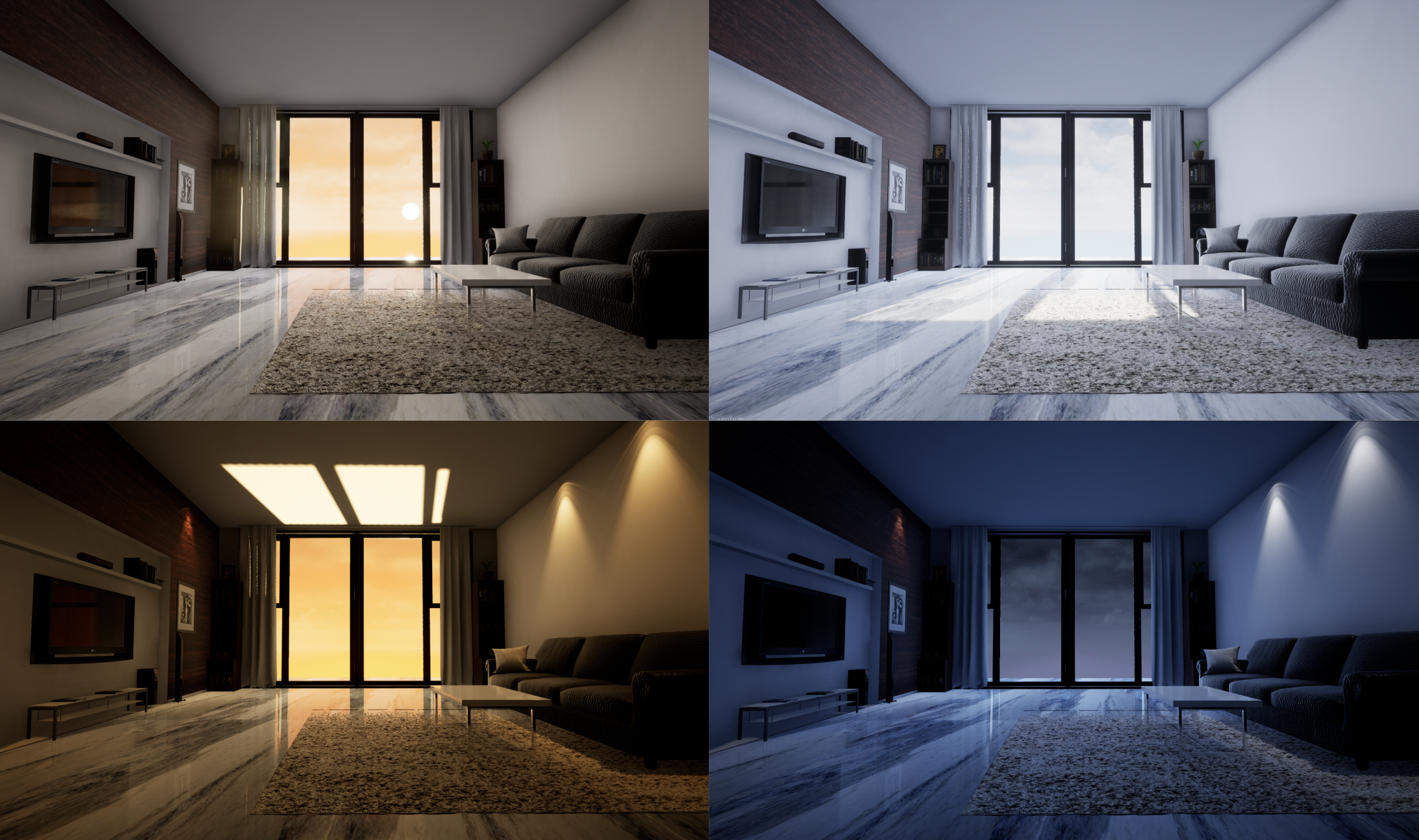}
\caption{Representative lighting conditions modeled in ElderSim (clockwise from top left: \textit{dawn}, \textit{noon}, \textit{night}, and \textit{sunset}). Controllable lighting conditions appropriately reflect the real-world with the usage of indoor light sources and the Post-Process Volume effect.}
\label{fig:lighting}
\end{figure}

    Lighting conditions in ElderSim are affected by both sunlight and indoor light sources modeled in UE4. To simulate the effect of sunlight over time, we utilize the SkySphere Blueprint function of UE4 and provide an adjustable time parameter in 100 levels to vary sunlight. Indoor light sources are placed according to lighting layouts of actual houses considered in Section~\ref{sec:background} and controlled to resemble our daily life better, e.g., turned off during the daytime and turned on during the evening as shown in Fig.~\ref{fig:lighting}. Rendering effects, which are significantly affected by lighting conditions, become finer by applying the Post-Process Volume effect of UE4.

\subsection{Object}

    Among the 55 activity classes considered in ElderSim, 35 activities contain human-object interaction. We model objects that are required to simulate these activities in UE4. The types of objects range from a single rigid body (for 28 classes) such as a cup to articulated objects such as a vacuum cleaner (for the \textit{vacuum the floor} class) or even deformable objects such as a jacket (for the \textit{take off jacket} class). All the objects are modeled in three different ways to increase diversity. When objects are used in ElderSim, they are attached to the contacting body parts' mesh and move along with the body parts to look natural. 
        
\subsection{User Interface}

\begin{figure}
    \centering\includegraphics[ width=0.5\linewidth, height=\linewidth, keepaspectratio]{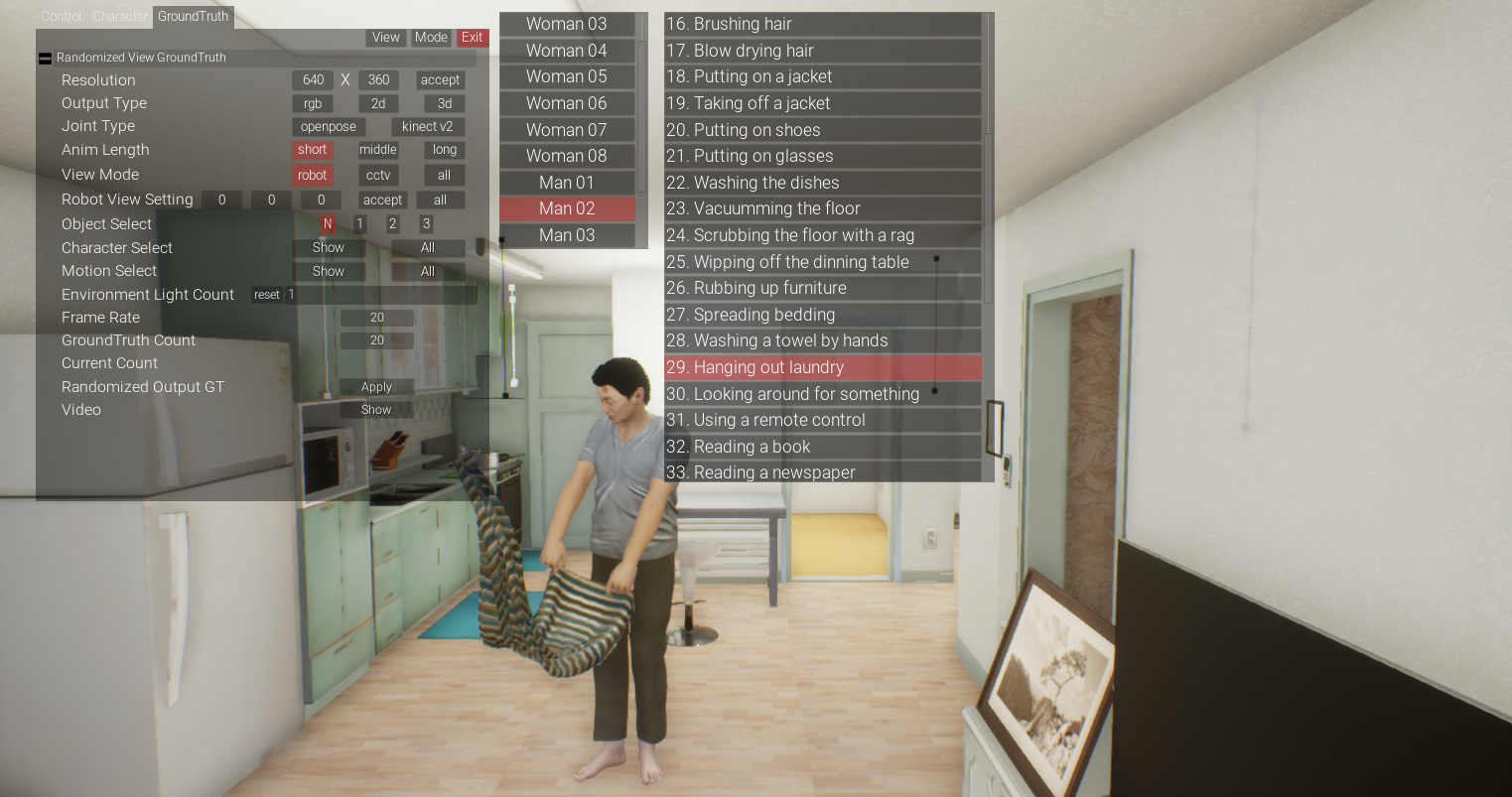}
    \caption{A snapshot of ElderSim showing the user interface to generate action data (a \textit{male elder} character performing \textit{hang out laundry}). The intuitive user interface allows the user to generate customized data by providing several data-generating options.}
    \label{fig:user_interface}
\end{figure}

\begin{figure}
    \centering\includegraphics[width=0.5\linewidth]{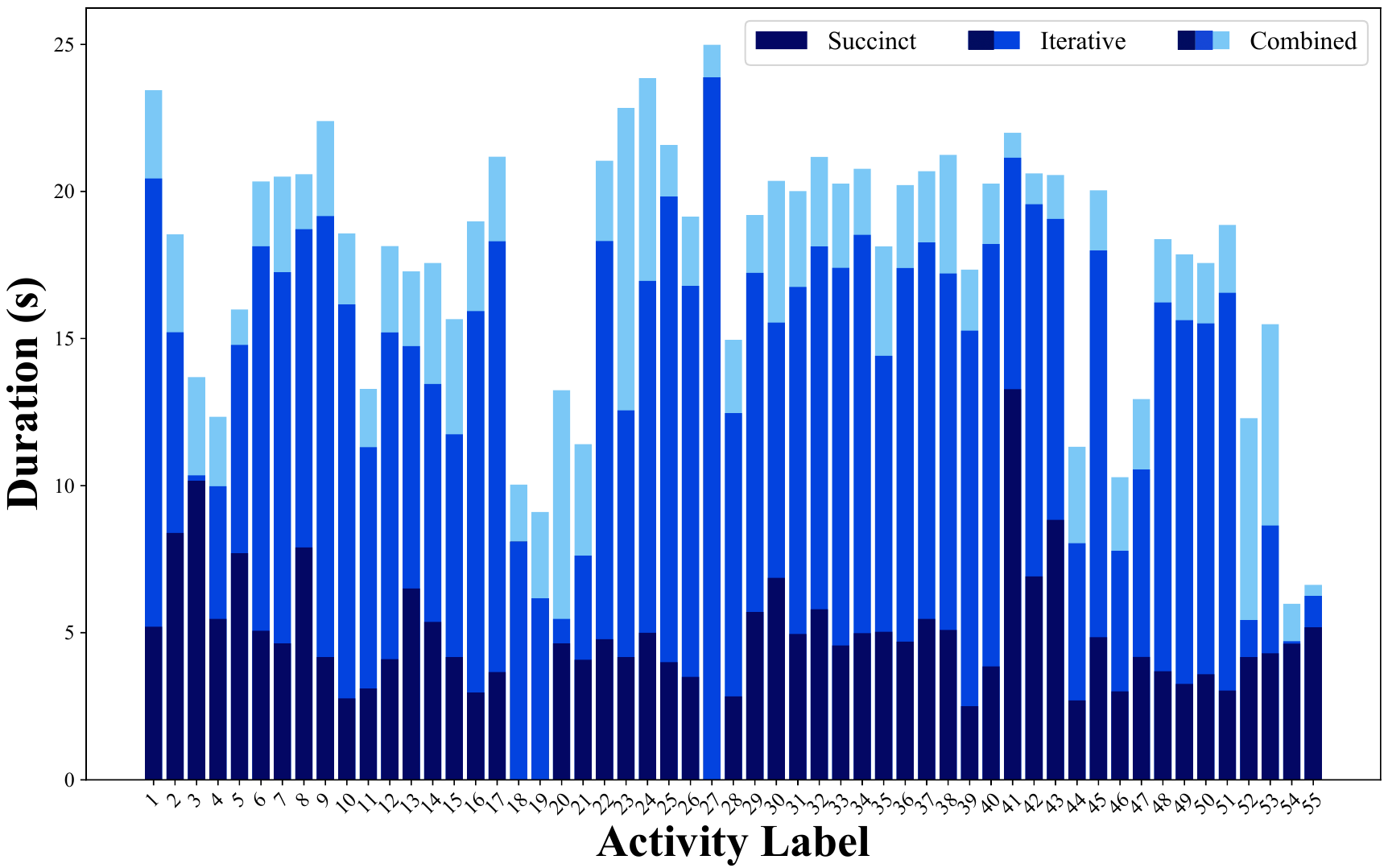}
    \caption{We provide three duration types in ElderSim to enhance applicability. The average video duration for each activity class provided in ElderSim is represented in different colors based on the duration types. Actions with a {\it Succinct} duration type contain a single trial of an action. {\it Iterative} duration-typed actions are performed repeatedly, but in a different way. Trivial motions are added to the {\it Iterative} duration type to form a {\it Combined} duration.} 
    \label{fig:durations}
\end{figure}

    An intuitive graphical user interface (GUI) is provided in ElderSim to select data-generating options as needed. The user can easily choose the desired subset of activities, characters, and backgrounds from the provided sets in order (see Fig.~\ref{fig:user_interface}). The camera viewpoints can then be selected among the robot- and surveillance-viewpoints, while preferable robot-viewpoints are adjusted by an angle interval $\phi$ and a set of heights $\delta$ as mentioned in Section~\ref{sec:viewpoint}. The lighting conditions are determined by choosing a subset of the hundred time-levels to vary sunlight, from 0 to 1. For the activities containing object interactions, the user can choose whether to use an object and which object model to use. In addition, for the activities containing repetitive motion, we provide three different types of motion duration to include one iteration (succinct), multiple iterations (iterative), and sequential movements (combined); the average motion duration for each activity class in ElderSim is illustrated in Fig.~\ref{fig:durations}.
    Once the data-generating options are determined, the data are automatically generated and recorded according to all possible combinations of options in ElderSim. ElderSim provides adjustable video resolutions and frame rates of up to $1920\times1080$ and 60 frames per second (FPS), respectively. Furthermore, three kinds of output data modalities are provided: RGB video, 2D, and 3D skeleton. For skeleton data, we provide both OpenPose- and Kinect v2-based skeletal formats. Parallel processing allows faster data generation.

\subsection{KIST SynADL}

    Based on the developmental features of ElderSim, we generate KIST SynADL, a large-scale synthetic dataset of elders’ daily activities considering care robot and smart surveillance applications. All 55 activities, 15 characters, and four backgrounds modeled in ElderSim are utilized to generate KIST SynADL. We further customize parameters for camera viewpoints as follows. To provide robot-viewpoints, we set the angle interval and height parameters (introduced in Section~\ref{sec:viewpoint}) to ${\{\phi=36^{\circ}}$, ${\delta=(\SI{0.7}{\metre}, \SI{1.2}{\metre})\}}$ 
    for semicircular camera layout respectively, where the height parameters are set based on several real-world care robots. These parameter settings result in ten horizontal camera locations, each having two different heights, thus providing 20 robot-viewpoints. Including eight more surveillance-viewpoints introduced in Section~\ref{sec:viewpoint}, KIST SynADL contains 28 camera viewpoints. For the lighting conditions, we divide a day into five parts by setting the time parameter to 0 (\textit{dawn}), 0.25 (\textit{noon}), 0.5 (\textit{evening}), 0.75 (\textit{sunset}), and 1 (\textit{night}), with \textit{noon} being the default lighting condition.  In the case of activities involving human-object interactions, we utilize only one kind of object model for each activity to generate data. RGB videos in the KIST SynADL dataset are recorded with a $640\times360$ resolution at 20 FPS, and corresponding 2D and 3D skeleton data are saved in both OpenPose- and Kinect v2-based formats. As a result, KIST SynADL provides 462k RGB videos, 2D, and 3D skeleton data, covering 55 action classes, 28 camera viewpoints, 15 characters, five lighting conditions, and four backgrounds.

\section{Experiments}
\label{sec:experiments}
    In this section, we experimentally validate and discuss the effect of augmenting our synthetic data, KIST SynADL (KIST) for training algorithms to recognize elders' ADL. We begin by introducing two real-world datasets for the experiments and address how insufficient the existing public dataset (NTU RGB+D 120) is to cover the elders' ADL. We then describe three state-of-the-art HAR methods used in the experiments as well as several experimental scenarios to examine the various aspects arising from the recognition of the elders' ADL. Within each experimental scenario, we investigate how our synthetic data can help recognize elders' daily activities and offer some guidance and insights for effective utilization of synthetic data.

\subsection{Datasets}
\label{sec:datasets}    
\begin{figure*}
    \centering\begin{tabularx}{0.8\textwidth}{@{}X*{4}{C}@{}}
        \centering\small{1. eat} &
        \textcolor{violet}{\fbox{\includegraphics[ width=\linewidth, height=\linewidth, keepaspectratio]{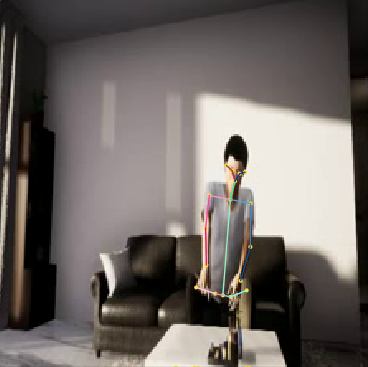}}} &
        \textcolor{pink}{\fbox{\includegraphics[ width=\linewidth, height=\linewidth, keepaspectratio]{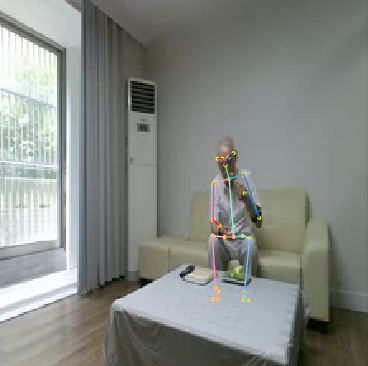}}} &
        \textcolor{cyan}{\fbox{\includegraphics[ width=\linewidth, height=\linewidth, keepaspectratio]{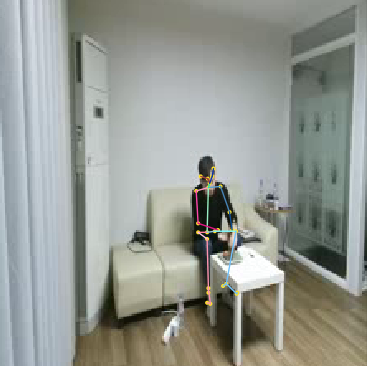}}} &
        \textcolor{cyan}{\fbox{\includegraphics[ width=\linewidth, height=\linewidth, keepaspectratio]{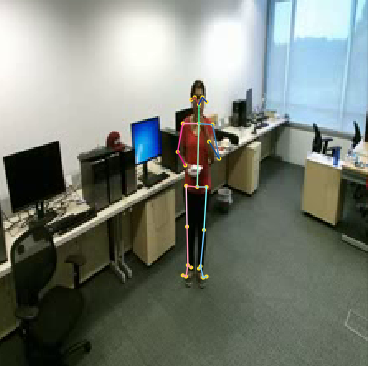}}}\\
        \centering\small{2. drink} &
        \textcolor{violet}{\fbox{\includegraphics[ width=\linewidth, height=\linewidth, keepaspectratio]{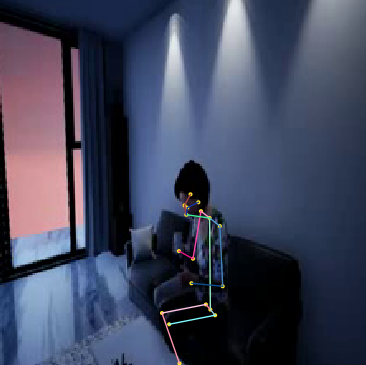}}} &
        \textcolor{pink}{\fbox{\includegraphics[ width=\linewidth, height=\linewidth, keepaspectratio]{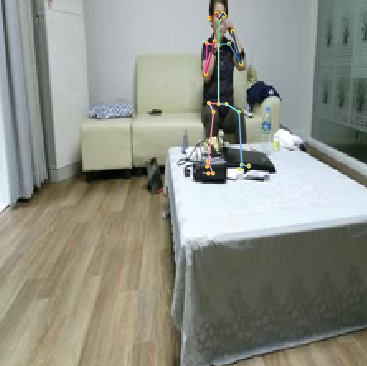}}} &
        \textcolor{cyan}{\fbox{\includegraphics[ width=\linewidth, height=\linewidth, keepaspectratio]{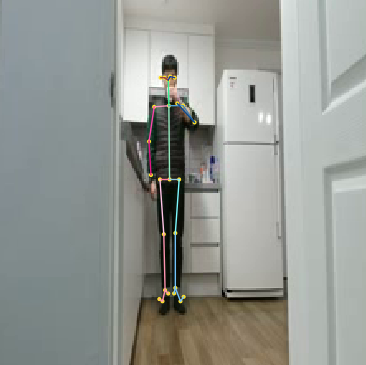}}}&
        \textcolor{cyan}{\fbox{\includegraphics[ width=\linewidth, height=\linewidth, keepaspectratio]{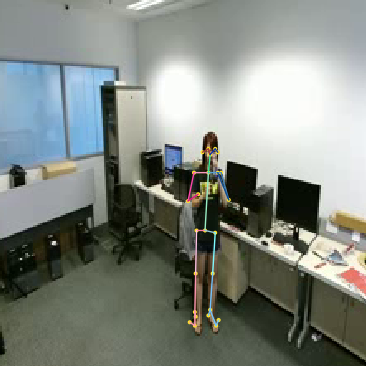}}} \\
        \centering\small{13. phone call} &
        \textcolor{violet}{\fbox{\includegraphics[ width=\linewidth, height=\linewidth, keepaspectratio]{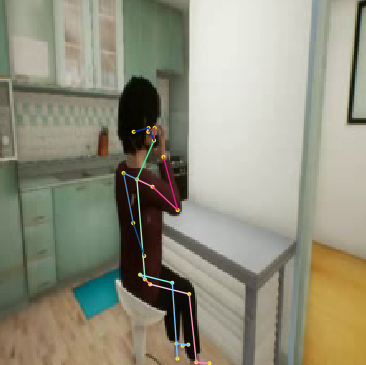}}} &
        \textcolor{pink}{\fbox{\includegraphics[ width=\linewidth, height=\linewidth, keepaspectratio]{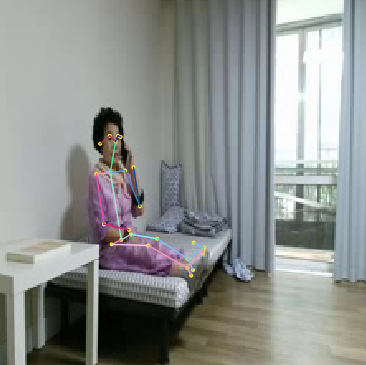}}} &
        \textcolor{cyan}{\fbox{\includegraphics[ width=\linewidth, height=\linewidth, keepaspectratio]{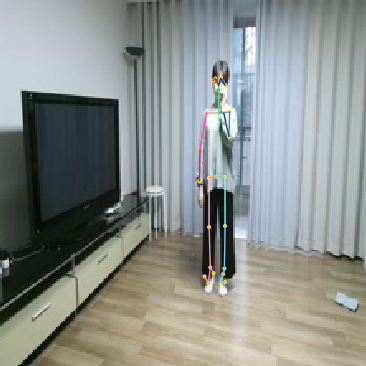}}}&
        \textcolor{cyan}{\fbox{\includegraphics[ width=\linewidth, height=\linewidth, keepaspectratio]{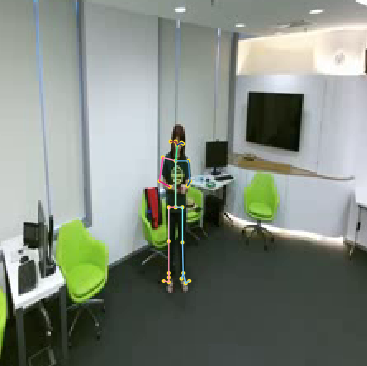}}} \\
        \centering\small{14. use computer} &
        \textcolor{violet}{\fbox{\includegraphics[ width=\linewidth, height=\linewidth, keepaspectratio]{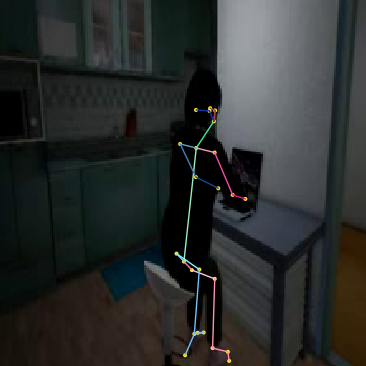}}} &
        \textcolor{pink}{\fbox{\includegraphics[ width=\linewidth, height=\linewidth, keepaspectratio]{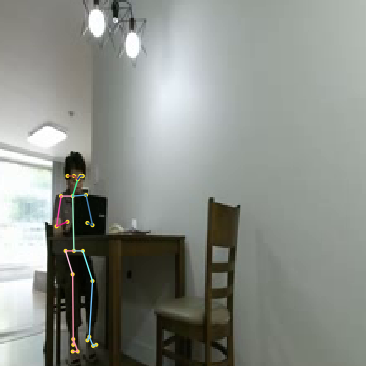}}} &
        \textcolor{cyan}{\fbox{\includegraphics[ width=\linewidth, height=\linewidth, keepaspectratio]{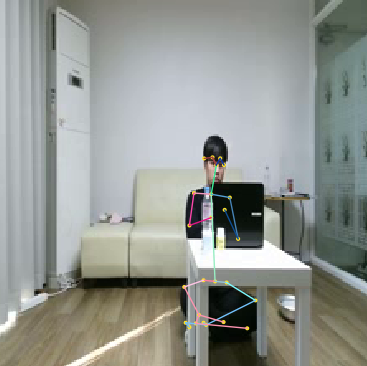}}} &
        \textcolor{cyan}{\fbox{\includegraphics[ width=\linewidth, height=\linewidth, keepaspectratio]{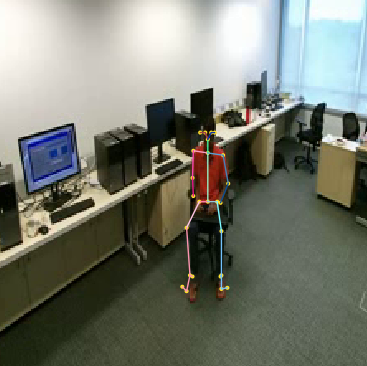}}}\\
        &\textbf{\small{KIST (Ours)}} &\textbf{\small{ETRI$_{E}$ [24]}} &\textbf{\small{ETRI$_{Y}$ [24]}} 
        &\textbf{\small{NTU [22]}} \\
    \end{tabularx}
\caption{Sample RGB snapshots and 2D skeleton coordinates of the datasets used in the experiments. The border color of each sample indicates both data type and age group (purple: synthetic-elderly data, pink: real-elderly data, cyan: real-young data).} 
\label{fig:datasets}
\end{figure*}

    We now introduce real datasets used in the experiments and explain how their activity classes are selected to match the elders' ADL. Samples of the datasets are visualized in Fig.~\ref{fig:datasets}.

\setlength{\fboxrule}{1.5pt}
\setlength{\fboxsep}{0pt}

\subsubsection{ETRI-Activity3D Dataset} 

     As introduced in Section~\ref{sec:related_Work}, the ETRI-Activity3D (ETRI) [24] dataset has been recently released for care robot applications to recognize the ADL of the elders'. This dataset contains 55 activity classes performed by a hundred subjects (composed of 50 elder and 50 young subjects) and captured from eight robot-viewpoints (installed in four locations each having two different heights) under constant lighting conditions. Since the dataset contains a large number of data with 112,564 samples, we utilize the ETRI dataset as our primary real-world dataset to train and evaluate three HAR models on elders' ADL.
    
\subsubsection{NTU RGB+D 120 Dataset} 

    The NTU RGB+D 120 [22] dataset consists of 120 kinds of activities performed by relatively young subjects in a laboratory environment. This dataset includes multiple trials of actions captured from five camera viewpoints installed at a constant height and under a constant lighting condition. Among 120 activities of [22], we construct the NTU dataset by selecting only 25 activities with 23,436 samples that match the 55 frequent ADL considered in the ETRI dataset. Such activities of the NTU dataset are mapped to the 23 activities of the ETRI dataset as shown in Table~\ref{table:dataset}. Even though [22] is one of the most widely investigated datasets in the HAR literature due to its large scale and diverse activity classes, there is still a severe gap in the subjects’ ages, backgrounds, and activity classes compared to the elders' ADL held in households. Therefore models trained on this dataset may not generalize well to the elders' ADL; we discuss such a point in the cross-dataset split with further details in Section~\ref{sec:experimental_scenarios}.

\begin{table*}
\begin{threeparttable}
\caption{Combined activity categories of the KIST, ETRI, and NTU datasets for the cross-dataset split. In some cases, more than two classes from NTU are merged to match a single class label in other datasets.}
\label{table:dataset}
\setlength\tabcolsep{0pt} 
    \begin{tabularx}{\linewidth}{@{\extracolsep{\fill}} XXXX}
        \toprule
             &\multicolumn{3}{c}{Dataset}\\
        \cmidrule{2-4}
        Combined Label &\textbf{KIST (Ours)} &\textbf{ETRI [24]} &\textbf{NTU [22]}\\
        \midrule
         1. eat    &eat food with a fork &eat food with a fork &eat meal\\
         2. drink  &drink water    &drink water      &drink water  \\
         3. brush teeth    &brush teeth  &brush teeth &brush teeth \\
         4. wash hands    &wash hands  &wash hands  &rub two hands\\
         5. brush hair  &brush hair   &brush hair   &brush hair \\
         6. wear clothes  &put on jacket  &put on jacket &put on jacket  \\
         7. take off clothes   &take off jacket &take off jacket &take off jacket \\
         8. put on/take off shoes  &put on/take off shoes     &put on/take off shoes      &put on+take off a shoe \\
         9. put on/take off glasses  &put on/take off glasses       &put on/take off glasses   &put on+take off glasses\\
         10. read    &read a book &read a book &read  \\
         11. write    &handwrite &handwrite &write\\
         12. phone call    &talk on the phone  &talk on the phone &phone call \\
         13. play with phone   &play with a mobile phone &play with a mobile phone &play with phone/tablet\\
         14. use computer    &use a computer &use a computer &type on a keyboard  \\
         15. clap    &clap  &clap &clap\\
         16. rub face    &rub face with hands &rub face with hands &wipe face \\
         17. bow    &take a bow  &take a bow &nod head/bow \\
         18. handshake    &handshake &handshake &shake hands \\
         19. hug    &hug each other &hug each other &hug  \\
         20. fight    &fight each other  &fight each other &punch/slap \\
         21. hand wave    &wave a hand &wave a hand &hand wave \\
         22. point finger    &point with a finger  &point with a finger &point to something  \\
         23. fall down    &fall down &fallen on the floor &fall down  \\
        \bottomrule
    \end{tabularx}
\smallskip
\scriptsize
\end{threeparttable}
\end{table*}

\subsection{Experimental Scenarios}
\label{sec:experimental_scenarios}
    We now introduce several experimental scenarios considered in the experiments. We begin by explaining {\bf cross-subject} and cross-view splits which are widely considered in the literature, and then introduce newly suggested cross-age and cross-dataset splits that assume the real and synthetic training datasets of different configurations.

    In the cross-subject or {\bf cross-view} splits, it is assumed that the real-world training dataset has limitations on available camera subjects or viewpoints. We train models using data acquired from only a part of the available subjects or viewpoints of a dataset and leave the remaining data as the test set. In the cross-subject split, we train the models on data for 24 subjects of the ETRI dataset and test on the other 76 subjects. In our cross-view split, we train on data for the two viewpoints (the seventh and eighth viewpoints of [24]) of the ETRI dataset and test on the other six viewpoints. 

    As an extension of the cross-subject split, we assume the situation in which the training data are only limited to one age group while having the other age group for evaluation. In such {\bf cross-age} splits, we divide the ETRI dataset into two subject groups of different ages, 50 younger subjects with the average age of 23.6 (ETRI$_{Y}$) and 50 elder subjects with the average age of 77.1 (ETRI$_{E}$). We then train the models on ETRI$_{Y}$ and test on ETRI$_{E}$ and vice versa. From such a split, we investigate if there are some differences according to the age groups in the recognition performance as well as the effect of utilizing our synthetic data.

    We further assume an extreme scenario of training a model to recognize ADL of the elders, while available data are far from those in several aspects. For example, data may be obtained only from young subjects in laboratory environments (e.g., the NTU dataset). In the {\bf cross-dataset} split assuming this scenario, we train models on the NTU dataset and test on the ETRI dataset, which corresponds to the dataset on elder's ADL. We then examine whether the models trained on the NTU dataset can be generalized well to the ETRI dataset and see if augmenting our synthetic data during training can help the generalization. Since the class composition of the datasets does not completely match each other, we define 23 combined classes based on the ETRI dataset and consider only 25 activity classes out of 120 for the NTU dataset to match the classes as explained in Section~\ref{sec:datasets} (see Table~\ref{table:dataset}).
    
    We then discuss the effect of training our synthetic KIST SynADL dataset for each scenario. From the baseline models trained without the KIST SynADL dataset, we investigate how action recognition performance varies when our synthetic dataset is augmented in the model training process. Furthermore, we vary the composition of the data used among the KIST SynADL dataset according to the scenarios. For example, since the ETRI dataset does not contain surveillance-viewpoints and diverse lighting conditions, we may use only robot-viewpoints and a default lighting condition of the KIST SynADL dataset when it is known to be tested on the ETRI dataset. For the later experiments, we use the abbreviation KIST for the KIST SynADL dataset containing only the default lighting condition. The experimental scenarios performed in this work is listed in Table~\ref{table:experiment_splits} with the variation factors that differ between training and test datasets.

\begin{table}[ht]
    \begin{threeparttable}
    \caption{The experimental scenarios and the factors that differ between training and test datasets for each split.}
    \label{table:experiment_splits}
    \setlength\tabcolsep{0pt}
        \begin{tabular*}{\linewidth}{@{\extracolsep{\fill}} lcccc}
            \toprule
            Variation Factor      &Dataset &Subject &View   &Age    \\
            \midrule
            Cross-Subject          &\xmark  &\large{\cmark}  &\xmark &\xmark  \\
            Cross-View             &\xmark  &\xmark  &\large{\cmark} &\xmark \\
            Cross-Age              &\xmark  &\large{\cmark}  &\xmark &\large{\cmark} \\
            Cross-Dataset          &\large{\cmark}  &\large{\cmark}  &\large{\cmark} &\large{\cmark} \\
            \bottomrule
        \end{tabular*}
    \smallskip
    \scriptsize
    
    \end{threeparttable}
\end{table}

\subsection{Training Details of HAR Methods}
    
    This section provides training details for the HAR methods utilized in the experiments, namely Glimpse Clouds (Glimpse) [11], Spatial Temporal Graph Convolution Network (ST-GCN) [12], and View Adaptive Convolutional Neural Network (VA-CNN) [13]. 

    Glimpse [11] is a RGB-based model that uses a visual attention module over the spatio-temporal cube to generate a cloud of glimpse windows. These windows are then soft-assigned to a set of gated recurrent units (GRUs) [54] that track the windows and process classification. A loss function to appropriately locate the windows is added to the original cross-entropy loss. Here, we follow [11] and utilize the 2D skeleton data corresponding to the RGB data to encourage the training process with another loss term that helps the model to perform pose regression. The Adam optimizer is used in training with an initial learning rate of 1e-4. Training the whole architecture took 13 hours for ten epochs with a minibatch size of 32 using a single NVIDIA Tesla V100 PCIe GPU. During test time, only RGB data resized to a $256\times256$ resolution is used as an input. We sample eight frames from a video sequence as in [50] and extract three windows per frame as inputs for the recurrent units.
    
    ST-GCN [12] represents 2D or 3D skeleton joint trajectories as a graph that connects nearby joints in a single frame and identical joints between consecutive frames. It then applies spatial temporal graph convolution on the constructed graph and captures the interaction between nearby joint groups and the temporal motions to facilitate action recognition. In this experiment, we apply ST-GCN on the 2D skeleton data. The 2D skeleton data of the real datasets are estimated from RGB videos using OpenPose, and pixel coordinates with the estimation confidence values of each joint are used as an input. We use the stochastic gradient descent to train ST-GCN models with batch size 64 for 50 epochs. The learning rates start at 0.1 and are reduced by 10 in epochs 20, 30, and 40. Moreover, when synthetic data is augmented in training, we split the last fully connected layer of the model so that the real and synthetic data can pass through different classifiers (except for the cross-dataset split). In this way the model empirically shows slightly better performance.

    VA-CNN [14] represents 3D skeleton joint trajectories as a planar image by mapping joint index and time axes to height and width axes respectively, and recognizes the image using a convolutional neural network (CNN). The most distinctive feature of the method is that it adapts the input data view to enhance the recognition performance with a view adaptation subnetwork. The subnetwork used to adapt the view is also modeled using a CNN, and the whole model is trained in an end-to-end fashion. We utilize Adam optimizer to train VA-CNN with batch size 64 for 30 epochs. The learning rate starts at 1e-4 and reduces by 10 for every ten epochs. We use the Kinect v2-based format for the KIST SynADL dataset to match real datasets.

    For the experiments augmenting the KIST SynADL dataset to train ST-GCN and VA-CNN, we balance mini-batches to contain an equal amount of real and synthetic data. Since the sizes of datasets differ, we randomly upsample the dataset of a smaller amount (usually the real-world data) to match the size. Twenty viewpoints of the KIST SynADL dataset were utilized for both methods, while only eight viewpoints were used for the Glimpse method to ensure reasonable training time. 
    
 \subsection{Experimental Results}
 
    We now report the results of the experiments performed according to the above settings. In the experiments, we trained three recognition algorithms for the proposed experimental splits and report the average video sequence-level top-1 classification accuracy for the five test trials as the action recognition score. For the results obtained from augmenting the KIST SynADL dataset, we designate the change in the recognition score from that obtained without augmentation in the parenthesis next to the score.
    
\subsubsection{Cross-Subject} 
    
\begin{figure*}
\includegraphics[width=\linewidth]{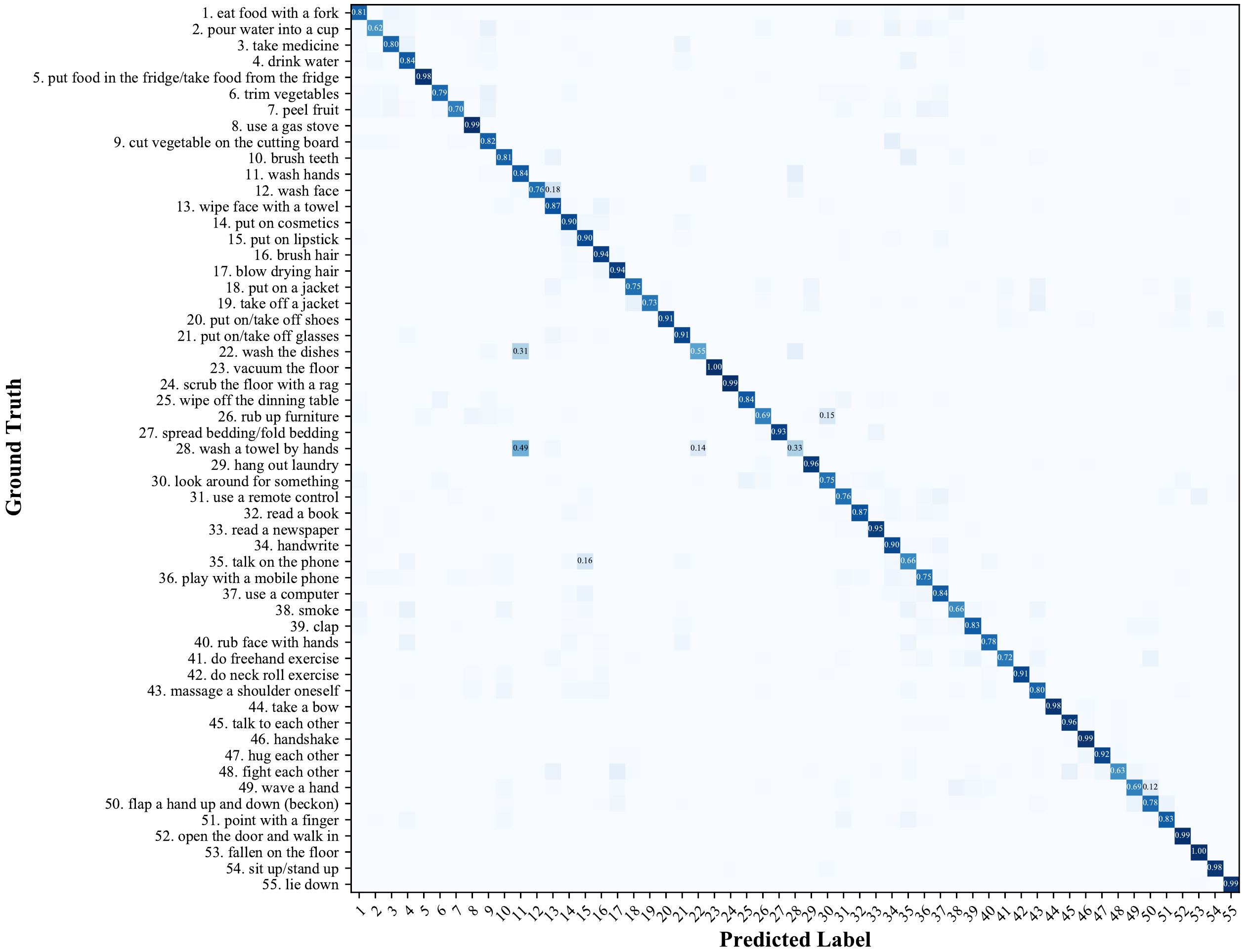}
\caption{Normalized confusion matrix of the Glimpse method trained by augmenting synthetic data in the cross-subject split. Only the values over 0.1 are displayed for better visualization.}
\label{fig:confusion_matrix}
\end{figure*}

    In the cross-subject split, 24 subjects (26,612 samples) from the ETRI dataset is sampled for the training set and evaluated by the remaining 76 subjects (85,912 samples) as explained in Section~\ref{sec:experimental_scenarios}. By augmenting synthetic data (26,400 samples for Glimpse and 66,000 samples for ST-GCN and VA-CNN) in training, each method's performance increases by 3.31, 0.68, and 0.22 percent points as described in Table~\ref{table:cross-subject}. While Glimpse shows the largest improvement, the absolute classification accuracy score is still lower than ST-GCN, showing confusion within some classes (e.g., data from {\it wash a towel by hands} class was frequently misclassified as {\it wash hands} class) as illustrated in Fig.~\ref{fig:confusion_matrix}. One should note that ST-GCN outperforms other methods in this cross-subject split of the ETRI dataset, while it performs worse than other considered methods on the NTU RGB+D [21] cross-subject split in the corresponding literature [11]--[13].

\begin{table}[ht]
    \begin{threeparttable}
    \caption{Accuracy comparison for the cross-subject split.}
    \label{table:cross-subject}
    \setlength\tabcolsep{0pt} 
        \begin{tabular*}{\columnwidth}{@{\extracolsep{\fill}} ccccc}
            \toprule
                 \multicolumn{2}{c}{Setting}  &\multicolumn{3}{c}{Top-1 Accuracy (\%)} \\ 
            \cmidrule{1-2}\cmidrule{3-5}
            Train     &Test &Glimpse [11] &ST-GCN [12] &VA-CNN [13] \\
            \midrule
            ETRI      &ETRI &80.22        &83.36       &81.98  \\ 
            ETRI+KIST &ETRI &\textbf{83.53 (+3.31)}    &\textbf{84.04 (+0.68)}  &\textbf{82.20 (+0.22)}  \\ 
            \bottomrule
        \end{tabular*}
    \smallskip
    \scriptsize
    \end{threeparttable}
\end{table}

\subsubsection{Cross-View}

    For the cross-view split, the data obtained from two camera viewpoints are used for training (26,757 samples) while the remaining six viewpoints (85,807 samples) are used for evaluation of the trained models. The cross-view results also demonstrate the benefit of training additional synthetic data for better performance as shown in Table~\ref{table:cross-view}. While ST-GCN shows the largest increase, Glimpse outperforms other methods which also differ from the results in the baseline studies [11]--[13]. Based on observation, this implies that the accuracies based on benchmark datasets do not perfectly correspond with the performance on another dataset (ETRI).
    
\begin{table}[ht]
    \begin{threeparttable}
    \caption{Accuracy comparison for the cross-view split.}
    \label{table:cross-view}
    \setlength\tabcolsep{0pt} 
        \begin{tabular*}{\columnwidth}{@{\extracolsep{\fill}} ccccc}
            \toprule
                 \multicolumn{2}{c}{Setting}  &\multicolumn{3}{c}{Top-1 Accuracy (\%)} \\ 
            \cmidrule{1-2}\cmidrule{3-5}
            Train &Test &Glimpse [11] &ST-GCN [12] &VA-CNN [13] \\
            \midrule
            ETRI               &ETRI &79.97          &77.88          &79.72  \\ 
            ETRI+KIST &ETRI &\textbf{81.59 (+1.62)}  &\textbf{80.84 (+2.96)}  &\textbf{80.00 (+0.28)} \\ 
            \bottomrule
        \end{tabular*}
    \smallskip
    \scriptsize
    \end{threeparttable}
\end{table}

\subsubsection{Cross-Age}

    In the cross-age split, we construct the training and test data from the ETRI dataset by splitting the data according to the subjects' age as explained in Section~\ref{sec:experimental_scenarios}, and examine if augmenting our KIST SynADL dataset in training affects the recognition performance differently according to the age group. The action recognition performances for the cross-age split experiments are shown in Table~\ref{table:cross-age}. Similarly to the previous results, synthetic data augmentation enhances recognition performance in most of the cases. By focusing on the performance change (the values placed in parentheses) induced from augmenting synthetic data, we observe that our synthetic data affect the recognition performance in a somewhat age-specific way, i.e., the augmentation seems more beneficial for the models trained on ETRI$_{Y}$ (and tested on ETRI$_{E}$) rather than those trained on ETRI$_{E}$ (and tested on ETRI$_{Y}$).     This effect is the most evident for the Glimpse method, which has the highest performance gain among the considered methods for the models trained on ETRI$_{Y}$ and even shows a performance decrease for the models trained on ETRI$_{E}$. Another interesting point to note is that, as can be observed in Table~\ref{table:cross-age}, the actions in ETRI$_{Y}$ seem to be more challenging to classify than the actions in ETRI$_{E}$ when the models are trained on the other data. This tendency agrees with the observation that the actions of the young subjects usually have larger motion differentials and shorter durations than the motions performed by the elders hence contain a wider variety [24]. Comparing the three HAR methods, ST-GCN outperforms other methods in the cross-age split. 
    
\begin{table}[ht]
    \begin{threeparttable}
    \caption{Accuracy comparison for the cross-age split.}
    \label{table:cross-age}
    \setlength\tabcolsep{0pt} 
        \begin{tabular*}{\columnwidth}{@{\extracolsep{\fill}} ccccc}
            \toprule
            \multicolumn{2}{c}{Setting}  &\multicolumn{3}{c}{Top-1 Accuracy (\%)} \\ 
            \cmidrule{1-2}\cmidrule{3-5}
            Train &Test &Glimpse [11] &ST-GCN [12] &VA-CNN [13] \\
            \midrule
            ETRI$_{E}$  &ETRI$_{Y}$ &\textbf{74.96} &77.52  &77.52\\ 
            ETRI$_{E}$+KIST &ETRI$_{Y}$ &73.90 (-1.06) &\textbf{78.12 (+0.60)} &\textbf{78.00 (+0.48)}\\
            \addlinespace[0.1cm]
            ETRI$_{Y}$  &ETRI$_{E}$ &75.35          &79.32          &78.06\\ 
            ETRI$_{Y}$+KIST &ETRI$_{E}$ &\textbf{77.74 (+2.41)}  &\textbf{80.38 (+1.06)}  &\textbf{78.18 (+0.12)}  \\ 
            \bottomrule
        \end{tabular*}
    \smallskip
    \scriptsize
    \end{threeparttable}
\end{table}

\subsubsection{Cross-Dataset}

    From the cross-dataset split, as explained in Section~\ref{sec:experimental_scenarios}, we examine whether a model trained on the NTU dataset (data from the young subjects in a laboratory background) can be generalized well to another (the ETRI dataset obtained from the elder subjects in daily-living environments) as well as the effect of augmenting our synthetic data during training. Table~\ref{table:cross-dataset} shows the result from the cross-dataset split tested on the ETRI dataset. The recognition performances for the cross-dataset split are lower than the results obtained from the former splits, in which the (real) training and test data come from a common dataset. These results imply that, for the eldercare services, it may not be sufficient to utilize deep models trained only on the NTU dataset, despite its large-scale.
    When synthetic data are augmented in training, we observe a firm performance increase for all the considered HAR methods for the cross-data split. The improvement gap is in general larger than the previous splits, with a remarkable boost for the Glimpse method (even over 13 percent point when tested on the ETRI dataset). Such a considerable increase in the Glimpse method may be partially because meaningful background information contained in RGB videos, which might be helpful to distinguish which activities are performed, is provided to the model from our synthetic data. In contrast, the NTU dataset alone might not provide much information on the backgrounds due to its limited laboratory setting. In Table~\ref{table:cross-dataset}, it is also interesting to observe that the recognition performance tested on ETRI$_{Y}$ is higher than that on ETRI$_{E}$; this may result from the fact that the NTU dataset contains actions of relatively young subjects.
    
\begin{table}[ht]
    \begin{threeparttable}
    \caption{Accuracy comparison for the cross-dataset split tested on the ETRI dataset.}
    \label{table:cross-dataset}
    \setlength\tabcolsep{0pt} 
        \begin{tabular*}{\columnwidth}{@{\extracolsep{\fill}} ccccc}
            \toprule
            \multicolumn{2}{c}{Setting}  &\multicolumn{3}{c}{Top-1 Accuracy (\%)} \\ 
            \cmidrule{1-2}\cmidrule{3-5}
            Train &Test &Glimpse [11] &ST-GCN [12] &VA-CNN [13] \\
            \midrule
            NTU         &ETRI &39.99         &46.92           &43.00  \\ 
            NTU+KIST    &ETRI &\textbf{54.79 (+14.80)} &\textbf{49.76 (+2.84)}  &\textbf{46.32 (+3.32)}  \\ 
            \addlinespace[0.1cm]
            NTU         &ETRI$_{E}$ &38.61         &45.66          &41.30   \\ 
            NTU+KIST    &ETRI$_{E}$ &\textbf{55.00 (+16.39)}  &\textbf{48.46 (+2.80)}  &\textbf{45.00 (+3.70)}  \\ 
            \addlinespace[0.1cm]
            NTU         &ETRI$_{Y}$ &41.18       &48.08           &44.58  \\ 
            NTU+KIST    &ETRI$_{Y}$ &\textbf{54.62 (+13.44)} &\textbf{50.92 (+2.84)}  &\textbf{47.48 (+2.90)} \\ 
            \bottomrule
        \end{tabular*}  
    \smallskip
    \scriptsize
    \end{threeparttable}
\end{table}
    
\subsubsection{Comparing VA-CNN and a Simpler Model Trained by Augmenting Synthetic Data}

    Using the dataset splits considered so far, we now propose a simple experiment to compare the effect of synthetic data augmentation to that of improving HAR neural network models. For ease of comparison, we adopt the VA-CNN model, in which the improvement of the neural network model is represented by implementing the view adaptation subnetwork (or VA module) [13]. Specifically, we compare the recognition performance of the baseline model, i.e., the VA-CNN without the VA module, trained by augmenting synthetic data and the VA-CNN method (the improved model from the baseline). According to the results in Table~\ref{table:VA-CNN_add}, the recognition performances from both settings are comparable to each other. In the cross-dataset split, augmenting synthetic data is superior to the model improvement. These results indicate that effective utilization of synthetic data during training can be a viable option for better HAR performance, as increasing the complexity of a neural network architecture.

\begin{table}[ht]
    \begin{threeparttable}
    \caption{Accuracy comparison between the VA-CNN method and the baseline model trained by augmenting synthetic data (odd rows for the former and even rows for the latter).}
    \label{table:VA-CNN_add}
    \setlength\tabcolsep{0pt} 
        \begin{tabular*}{\columnwidth}{@{\extracolsep{\fill}} ccccc}
            \toprule
            \multicolumn{4}{c}{Setting}  &Top-1 Accuracy (\%) \\ 
            \cmidrule{1-4}\cmidrule{5-5}
            Split &Train &Test                    &VA Module &VA-CNN [13]     \\
            \midrule
            \multirow{2}{*}{Cross-Subject} &\multirow{2}{*}{$\left\lbrace\begin{array}{c}
            \text{ETRI} \\
            \text{ETRI+}\text{KIST}\end{array}\right.$}             &ETRI          &\cmark  &81.98  \\ 
            &   &ETRI   &\xmark &\textbf{82.26 (+0.28)}  \\ 
            \addlinespace[0.2cm]
            \multirow{2}{*}{Cross-View}  &\multirow{2}{*}{$\left\lbrace\begin{array}{c}
            \text{ETRI} \\
            \text{ETRI+}\text{KIST}\end{array}\right.$} &ETRI &\cmark  &\textbf{79.72 (+0.04)}    \\ 
            &      &ETRI    &\xmark  &79.68  \\ 
            \addlinespace[0.2cm]
            \multirow{4}{*}{Cross-Age} &\multirow{4}{*}{$\left\lbrace\begin{array}{c}
            \text{ETRI$_{E}$} \\
            \text{ETRI$_{E}$+KIST} \\
            \addlinespace[0.1cm]
            \text{ETRI$_{Y}$} \\
            \text{ETRI$_{Y}$+KIST} \\
            \end{array}\right.$} &ETRI$_{Y}$    &\cmark         &\textbf{77.52 (+0.10)}       \\ 
            & &ETRI$_{Y}$    &\xmark &77.42  \\ 
            \addlinespace[0.1cm]
            & &ETRI$_{E}$    &\cmark &\textbf{78.06 (+0.18)} \\ 
            & &ETRI$_{E}$    &\xmark &77.88  \\ 
            \addlinespace[0.2cm]
            \multirow{6}{*}{Cross-Dataset}  &\multirow{6}{*}{$\left\lbrace\begin{array}{c}
            \text{NTU} \\
            \text{NTU+}\text{KIST} \\
            \addlinespace[0.1cm]
            \text{NTU} \\
            \text{NTU+}\text{KIST} \\
            \addlinespace[0.1cm]
            \text{NTU} \\
            \text{NTU+}\text{KIST}\end{array}\right.$}  &ETRI    &\cmark &43.00 \\ 
            &    &ETRI         &\xmark  &\textbf{44.94 (+1.94)}  \\ 
            \addlinespace[0.1cm]
            &    &ETRI$_{E}$   &\cmark  &41.30  \\ 
            &    &ETRI$_{E}$   &\xmark  &\textbf{43.70 (+2.40)}  \\ 
            \addlinespace[0.1cm]
            &    &ETRI$_{Y}$   &\cmark  &44.58  \\ 
            &    &ETRI$_{Y}$   &\xmark  &\textbf{46.10 (+1.52)}  \\  
            \bottomrule
        \end{tabular*}
    \smallskip
    \scriptsize
    \end{threeparttable}
\end{table}

\section{Conclusion}
\label{sec:conclusion}
    Considering eldercare applications, obtaining data of elders' activities of daily living is necessary, but challenging. We take advantage of modern realistic rendering and visualization techniques to develop a platform named ElderSim and simulate a variety of daily activities of the elderly. Based on ElderSim, we generate a large-scale synthetic dataset of elders' activities, KIST SynADL dataset, considering possible applications for care robots and smart surveillance systems. We then demonstrate the effectiveness of augmenting the KIST SynADL dataset in training from extensive experiments involving three state-of-the-art HAR methods as well as two different real-world human activity datasets. We show noticeable improvements of action recognition performance by augmenting our synthetic data. We also offer guidance and insights for the effective utilization of our synthetic data in human action recognition.

    In the future, we plan to enlarge the subject diversity by changing the body shape of the elderly and applying the corresponding motion styles to actions. Furthermore, we will extend our ElderSim platform by employing additional features of UnrealCV [55] to apply to more various problems in computer vision and robotics. Designing a domain-adaptive learning framework for HAR to further utilize our synthetic data would be another intriguing area of future research.

\section*{Acknowledgment}
\label{sec:Acknowledgment}
    The authors thank Jaewang Lee for his supportive work with realistic 3D modeling of implemented features in ElderSim.

\bibliographystyle{unsrt}

\end{document}